\documentclass{article}
\usepackage[a4paper,left=3cm,right=3cm,top=2cm, bottom=2cm]{geometry}

\usepackage{amsmath}
\usepackage[utf8]{inputenc} 
\usepackage[T1]{fontenc}    
\usepackage{hyperref}       
\usepackage{url}            
\usepackage{booktabs}       
\usepackage{amsfonts}       
\usepackage{nicefrac}       
\usepackage{microtype}      
\usepackage{subfig}
\usepackage{graphicx}
\usepackage{amssymb}
\usepackage{centernot}
\usepackage{tabularx}
\usepackage{makecell}
\newtheorem{theorem}{Theorem}[section]

\newtheorem{definition}[theorem]{Definition}

\newtheorem{remark}[theorem]{Remark}

\newenvironment{proof}{\paragraph{Proof:}}{\hfill$\square$}
\usepackage[ruled,vlined,linesnumbered,noresetcount]{algorithm2e}
\SetKwInOut{Parameter}{Parameters}
\title{Capacity Preserving Mapping  for High-dimensional Data Visualization}

\author{%
  Rongrong~Wang \quad 
 Xiaopeng Zhang }
   
\begin{document}
\date{}
\maketitle

\begin{abstract}
We provide a rigorous mathematical treatment to the crowding issue in data visualization when high dimensional data sets are projected down to low dimensions for visualization.  By properly adjusting the capacity of high dimensional balls, our method allows the high dimension data to be embedded into low dimensions without being crowded near a specific region.  A key component of the proposed method is an estimation of the correlation dimension at various scales which reflects the data density variation. The proposed adjustment to the capacity applies to  any distance (Euclidean, geodesic, diffusion) and can potentially be used  in many existing methods to mitigate the crowding issue during the dimension reduction.  We demonstrate the effectiveness of the new method using synthetic and real datasets.
\end{abstract}


\section{Introduction}

Visualizing high dimensional data via their low dimension projections is particularly useful in facilitating data analysts to understand their datasets, detecting underlying data patterns and creating various hypotheses about the data. To achieve this goal, we project the high dimensional dataset $X=\{X_1,...,X_N\} \subset \mathbb{R}^n$  down to low dimensions: $Y_i = P(X_i) \in \mathbb{R}^d$ ($i=1,...,N$) with $d=2$ or 3 and visualize the low dimensional embedding $\{Y_i\}_{i=1}^N$ via a single scatter plot. Designing the mapping $P$ that yields a reliable visualization is the focus of this paper.  A good design should take into account the special need of data visualization 1) as most human beings can only digest visual information in at most three dimensions (or four dimensions if video sequences are included) , the map is required to project data of any  dimensionality down to 2 or 3. 2)  as revealing the data pattern is the major objective, the map should be able to preserve the geometrical structure of the data. 

Most existing dimensionality reduction techniques (i.e., MDS \cite{torgerson1952multidimensional}, LLE \cite{roweis2000nonlinear}, Isomap \cite{tenenbaum2000global}) are only designed to reduce the data to its intrinsic dimension, which is usually higher than three.  While data visualization methods SNE \cite{hinton2002stochastic}, tNSE \cite{van2008visualizing}, UMAP \cite{mcinnes2018umap}, PHATE \cite{moon2019visualizing} can reduce data of any dimension to two or three, they usually cannot preserve geometric relations such as cluster radii and relative distances between clusters. We hereby formulate the main mathematical question  in data visualization: how to map datasets with a wide range of dimensionality to 2 or 3 while minimizing the geometric  distortion?

The geometric distortion we care about is the relative distance/similarity between points.  It is considered successfully preserved if points close to each other remain close after embedding and those far away from each other remain far away. The main obstacle in preserving the geometric relation is the so-called crowding issue. Simply put, the crowding issue arises from the fact that a higher dimensional body typically has a larger capacity than a lower dimensional one, hence reducing the dimensionality causes points to be crowded. 

We propose a way to adjust the capacity of the high dimensional body before the dimension reduction. The method named Capacity Preserving Mapping (CPM) is essentially a class of methods that generalize many existing methods by redefining the distance they use. Compared to the popular methods such as tSNE and UMAP, our method can better preserve geometrical structures of the dataset and does not presume existences of clusters.


We note that although previous methods SNE, t-NSE, UMAP and non-metric MDS also treat the crowding problem to some extent, the fundamental capacity mismatch between different dimensions, which is the main cause of the crowding problem, were not yet carefully analyzed or addressed. Our rigorous analysis of the capacity mismatch and the proposed way of adjusting it will help to avoid the crowding issue with a minimal amount of distortion.

\section{Related work}
Data visualization is an important task in data mining and is closely related to dimensionality reduction, graph learning, and data clustering.  A number of early works studied the so-called table data visualization problem (see the review article \cite{de2003visual}) that visualizes $N$ ($N>3$) attributes of data in a table using 2D or 3D plots. This is made possible either by using multiple pixels/attributes/coordinates in the low dimension to represent one high dimensional data point or using multiple plots from different angles to build up the high dimensional image. The drawbacks of these methods are that the visualization is not directly digestible, needs human effort to understand and interpret, and the relation between data points (such as whether clusters exist or how close they are) is not immediately apparent. 

Along a separate line of research, one aims to visualize high dimensional data in one scatter plot through various dimension reduction methods \cite{van2009dimensionality}.  The methods can be categorized as linear ones (PCA, MDS \cite{torgerson1952multidimensional}, ICA \cite{hyvarinen2000independent}, etc) and nonlinear ones (LLE \cite{roweis2000nonlinear}, non-metric MDS \cite{rabinowitz1975introduction}, Isomap \cite{tenenbaum2000global}, Laplacian Eigenmap \cite{belkin2001laplacian}, Diffusion map \cite{lafon2006diffusion}, etc). 
 As mentioned earlier,  these methods can only reliably map the data down to its intrinsic dimension, while visualization requires the target dimension to be less or equal to three. Hence these dimensionality reduction methods work really well on artificial datasets with small intrinsic dimensions (e.g., swiss roll (2D), Helix (1D), Twin peaks (2D)), but not as well on real datasets (e.g., MNIST ($\sim$ 10D), COIL($\sim$ 5D)).  
 
The class of methods that are most relevant to ours includes SNE \cite{hinton2002stochastic}, tSNE \cite{van2008visualizing}, and UMAP \cite{mcinnes2018umap} \footnote{There exist many other more domain-specific visualization techniques such as PHATE \cite{moon2019visualizing} that emphasizes on biological trajectory data.}. They are data visualization methods rather than dimensionality reduction methods, and the goal of these methods is to produce good visualization results in two or three dimensions no matter what the actual intrinsic dimension of the data is.  
However, these methods are designed to only preserve local (i.e., neighbourhood) information and will enforce the formation of clusters even when the original dataset does not contain any cluster.  
The purpose of this paper is to propose a  non-local geometry preserving algorithm that aims to preserve distances of all scales as much as possible, therefore serving as a good supplement to the existing clustering methods. Since the main obstacle of distance preservation is the capacity mismatch between different dimensions,  our focus will then be carefully computing the intrinsic dimensions at various scales and using them to adjust the capacity. 
\section{Motivation - the crowding phenomenon}\label{sec:3}
As stated in \cite{van2008visualizing}, when high dimensional data is mapped to low dimensions, there is a tendency for non-overlapping groups to overlap. Theoretically, this is due to the difference in norm concentration between high and low dimensions:  a ball in higher dimensions has a volume that grows faster with radius $\textrm{Vol}(B_2^n(r))\sim r^n$ \cite{falconer2004fractal}, where $B_2^n(r)$ is the $\ell_2$ ball in $\mathbb{R}^n$ with radius $r$. This is saying that when the dimension $n$ is large and data points are uniformly distributed in $B_2^n(r)$, one can find more points near the surface of $B_2^n(r)$ than those around the center. When all these points are mapped down to low dimensions,  since there is now less room near the surface, the points will be pushed towards the center, causing a distortion of the geometry.  

\begin{figure}
\centering
     \subfloat[][$n=2$]{\includegraphics[scale=0.2]{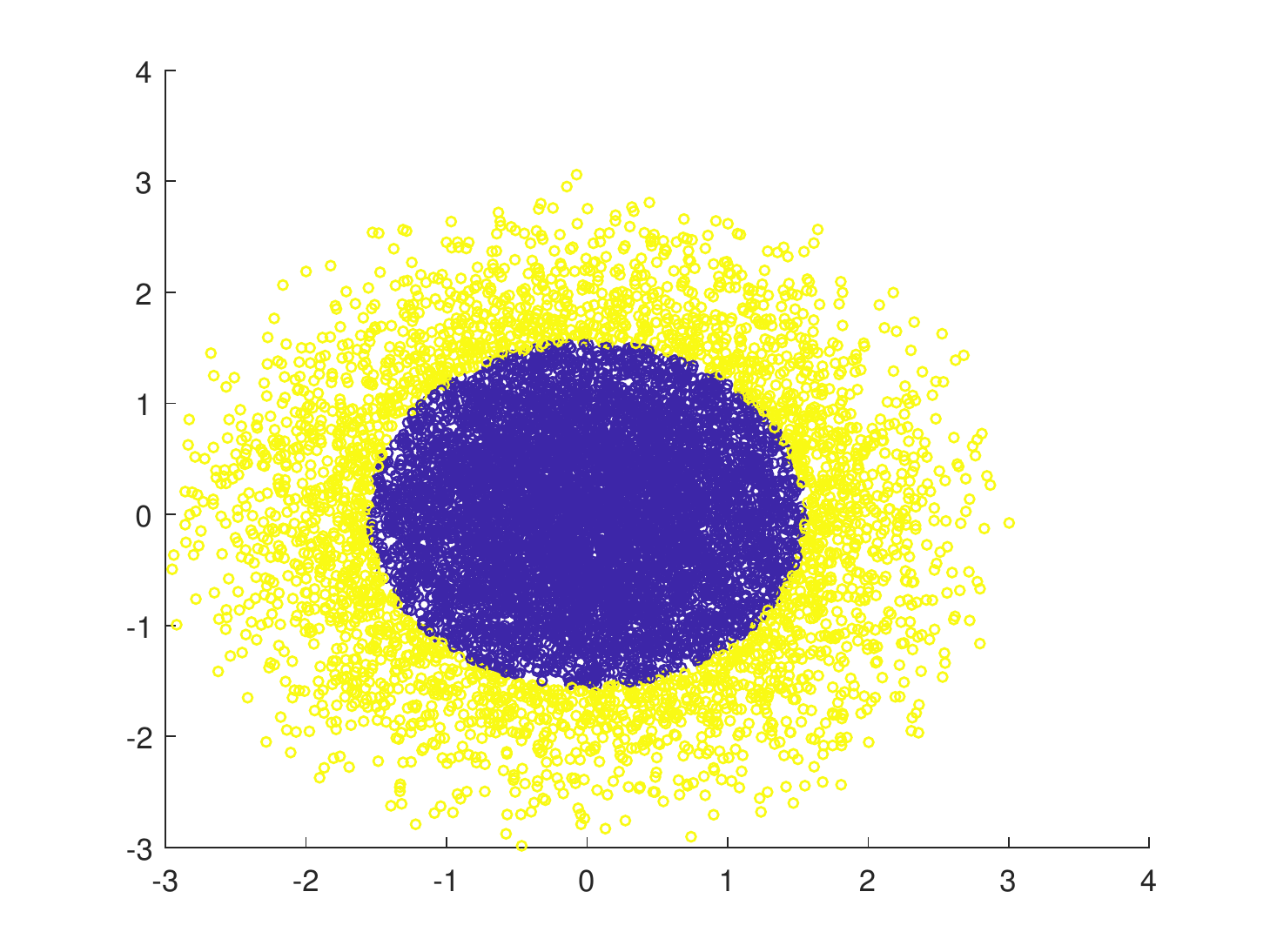}\label{figure1a}}
     \subfloat[][$n=5$]{\includegraphics[scale=0.2]{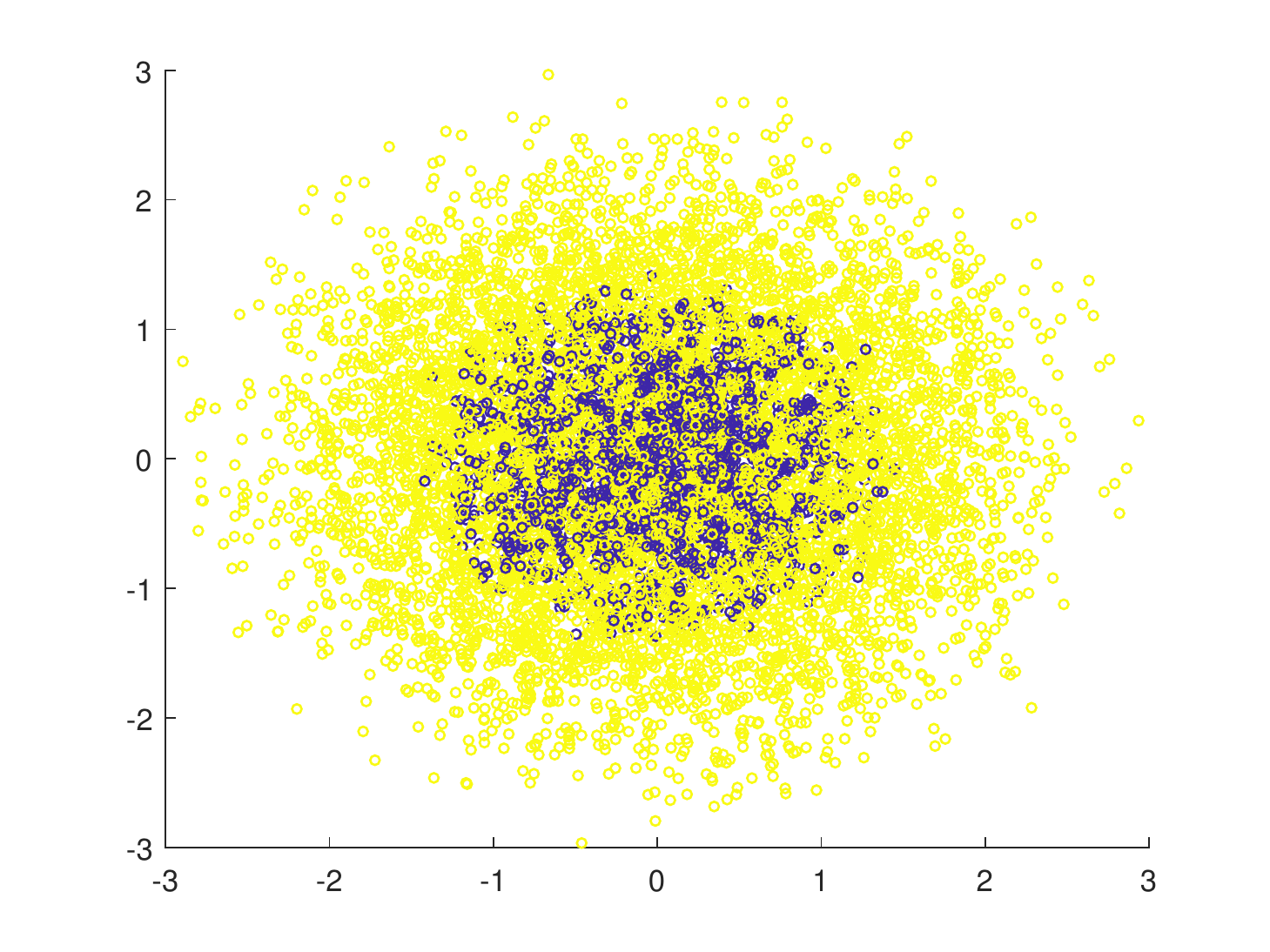}\label{figure1b}}
     \caption{The crowding phenomenon: two non-overlapping classes (a) becomes overlapping after MDS embedding (b)}
     \label{fig:motivating}
     \end{figure}

Let us visualize the crowding phenomenon observed during dimension reduction via  Multidimensional Scaling (MDS).   Assume a high dimensional $\ell_2$ ball contains two classes of points.  Class 1 lies inside the ball $B_2^n(1)$ and Class 2 lies in a spherical shell right outside Class 1.  When $\{X_i\}_{i=1}^N\subseteq \mathbb{R}^2$ is two dimensional, a direct visualization (Figure: \ref{figure1a}) shows the correct relation between the two classes. When $\{X_i\}_{i=1}^N\subseteq \mathbb{R}^5$ is five dimensional and is mapped down to $\mathbb{R}^2$ via MDS, its geometrical structure is distorted (Figure \ref{figure1b}) and we see a severe crowding phenomenon: the second class is pushed towards the center.  Finding a way to correct this type of distortion is our main objective. 
     \subsection{Notation}\label{sec:notation} 
     Let $f(x)\geq 0$ and $g(x)\geq 0$ be two increasing functions of $x$. We use $f\simeq g$ to denote that there exist constants $0<c<C$ such that $ cg(x) \leq f(x)\leq Cg(x)$ for all $x$. $f(x)\lesssim g(x)$ means $\frac{f'(x)}{f(x)}< \frac{g'(x)}{g(x)}$, i.e., $\log f(x)$ grows slower than $\log g(x)$ as $x$ increases, and $f(x)\gtrsim g(x)$  means the opposite. 
     For a random variable $z$ and its probability density function $h$, we use $P_{z\sim h}(z\in A)$ to denote the probability of the event $A$ under $h$. 
   
\section{The methodology}
\subsection{Characterizing the capacity of a manifold}
To mathematically describe the crowding issue, we need to rigorously define the ``capacity'' of regions of a manifold. Intuitively, the capacity of a region should reflect the amount of data the region holds. We are interested in both the static capacity (the amount of data held by a neighbourhood) and the dynamic capacity (the growth rate of the capacity as the radius of the neighbourhood increases). We hope to make the embedding preserve these capacities. 

Since the amount of data a region of the manifold holds not only depends on the volume of the region but also the sampling density, we assume that each data point $X_i$ is independently drawn from an underlying manifold $\mathcal{M}$ according to a continuous sampling probability distribution $f(\mathcal{M})$. 
\begin{definition}[Relative capacity]Let $\mathcal{M}$ be a compact manifold  equipped with a metric $\|\cdot\|$ and sampling distribution $f$: $\mathcal{M}\rightarrow \mathbb{R}$ .  For any point $p \in \mathcal{M}$, we define \textbf{the relative capacity of the radius $r$ neighbourhood of $p$}  as the probability that randomly drawn points according to $f$  fall inside this neighbourhood, 
\begin{equation}\label{eq:C}
C(p, r;\mathcal{M},f, \|\cdot\|) := \mathbb{P}_{z\sim f(\mathcal{M})} (\|z-p\|\leq r) = \mathbb{E}_{z \sim f(\mathcal{M})} \mathbf{I}(r-\|z-p\|))  
\end{equation}
where  $\mathbf{I}$ is the step function used to model the neighbourhood of radius $r$ centred on $p$ (i.e., $\{z: \|z-\|\leq r\})$. We define the \textbf{average capacity of the radius $r$ neighbourhood} as taking average of $C(p,r;\mathcal{M},f, D)$ cross all locations $p \in \mathcal{M}$ of the manifold,
\[
C(r;\mathcal{M},f,D) : = \mathbb{E}_{p\sim f(\mathcal{M})} C(p,r;\mathcal{M},f,D),
\]
For simplicity, we write $C(r), C(p,r)$ in short for $C(r;\mathcal{M},f,D)$, $C(p,r;\mathcal{M},f,D)$, respectively.
\end{definition} Intuitive, the ``capacity'' of a neighbourhood reflects the amount of data the neighbourhood holds. The ``relative capacity'' is the normalized capacity, and hence is a probability measure. If many samples are generated, the relative capacity $C(p,r)$ reflects the expected number of points falling in a neighbourhood. If the neighbourhood size $r$ is fixed, the relative capacity $C(p,r)$ as a function of $p$ reflects how the data density varies across different locations $p$ on the manifold. If the location $p$ is fixed, $C(p,r)$ as a function of $r$ reflects how fast the capacity grows as the neighbourhood expands.
The average capacity $C(r)$ for the manifold is the average of the pointwise one $C(p,r)$, and equals to the percentage of random pairs falling inside any radius $r$ neighbourhood on the manifold. 
Since by definition, $C(p,r)$ and $C(r)$ are cumulative distribution functions, their partial derivatives $\frac{\partial C(p,r)}{\partial r} $, $\frac{\partial C(r)}{\partial r} $ are probability density functions. 
\begin{definition}[Relative density] The relative density functions are defined as the derivatives of the relative capacities
\[\rho(p,r) := \frac{\partial C(p,r)}{\partial r},  \quad \rho(r) := \frac{d C(r)}{d r}\]
\end{definition}
We also have the reversed relation  $C(p,r)=\int_0^r \rho (p,t)dt$, and $ C(r)=\int_0^r \rho (t)dt$.

\textbf{Exmaple 1:} Suppose $\mathcal{M} = B^n_2(1)$, $f(\mathcal{M})$ is the uniform distribution on $\mathcal{M}$, and the metric $\|\cdot \|$ on $\mathcal{M}$ is the Euclidean distance. Then it is straightforward to verify that $C(0,r;\mathcal{M},D) = r^n$, for any $r\leq 1$, and $C(p,r;\mathcal{M},D) \simeq r^n$ where $\simeq$ was defined in Sect \ref{sec:notation}. Taking expectation with respect to $p$, one can verify that the capacity of the manifold is $C(r;\mathcal{M},D) \simeq r^n$ and the density is about $\rho(r) \simeq nr^{n-1}$.

In what follows, we make the following two assumptions on the data. \\
\textbf{Assumption 1}:  The original high-dimensional data $\{X_i\}_{i=1}^N$ are drawn independently from some underlying manifold $\mathcal{M}$ according to a continuous distribution $f(\mathcal{M})$, and the manifold $\mathcal{M}$ is equipped with a given metric $\|\cdot\|$.  The embedded data  $Y_i=P(X_i)$, $i=1,...,N$,  are independent realizations of the embedded manifold $\mathcal{S}=P(\mathcal{M})$, according to the induced distribution $f_I$ of $f$ under $P$. \\
For the visualization of the low dimensional embedding  to be easy to interpret, we insist the  low dimensional metric to be Euclidean and the low dimensional sampling distribution $f_I$ to be the uniform distribution with respect to this metric. That is to say, we hope to design an embedding map $P$ so that the induced distribution $f_I$ is the uniform Euclidean measure of $\mathbb{R}^d$ restricted to the embedded manifold $\mathcal{S}$. Hence in the following, when  computing the relative capacity/density for the low dimensional manifold we always use the uniform measure, $f_{uniform}$, and the usual $\ell_2$ metric $\|\cdot\|_2$. We emphasize that $\|\cdot\|_2$ is different from $\|\cdot\|$, where the latter represents the (arbitrarily) given metric for the high dimensional space. In practice, $\|\cdot\|$ can be chosen as Euclidean, geodesic or other distances between the pairwise data.  \\
\textbf{Assumption 2}: The relative densities $ \rho(r; \mathcal{M},f, \|\cdot \|) $ and $ \rho(r; \mathcal{S},f_{uniform}, \|\cdot \|) $ of the original manifold $\mathcal{M}$ and the embedded one $\mathcal{S}$  can be fitted with the models
\[ \rho(r; \mathcal{M},f, \|\cdot \|) = c_mn_m(r) r^{n_m(r)-1}, \quad \rho(r; \mathcal{S},f_{uniform}, \|\cdot \|_2)  = c_s n_s(r) r^{n_s(r)}\]
where $c_m$ and $c_s$ are absolute constants, and $n_m(r)>0$, $n_s(r)>0$ are both slowly varying dimension functions of $r$. 

In Assumption 2, we are essentially assuming that both densities are of the form $nr^{n-1}$ with $n$ being a function of the neighbourhood radius $r$. Let us first explain why the form $nr^{n-1}$ is imposed and then why $n$ has to change with $r$. Recall that for a ball with intrinsic dimension $n$, if  the data density is uniform in the ball, then the capacity of a radius $r$ neighourbood  is about $ r^n$ (Example 1). Thinking of this neighbourhood as consisting of concentric shells with  the same thickness and increasing radii, then under the uniform density assumption,  the shell with radius $r$ will hold $\sim nr^{n-1}$ data points. By definition, the relative density is the derivative of capacity of the shells as the thickness approaches 0, so they naturally inherit the form $nr^{n-1}$, where $n$ being the intrinsic dimension at scale $r$. Comparing to the uniform density assumption, a more realistic assumption is that the data density varies with the scale, which is equivalent to saying that the intrinsic dimension $n$ varies with the scale, thus we assumed $n$ to be a function of $r$ in Assumption 2.  Note that strictly speaking, $n$ also various with angles, but we drop that independence to simply the problem. 

\begin{remark}\label{rm:corr}
$n(r)$ and $n(s)$ are closely related to the so-called correlation dimension \cite{grassberger2004measuring}. Take $n(r)$ as an example, the correlation dimension is defined as $\textrm{dim}_{corr}=\lim\limits_{r\rightarrow 0} \frac{\partial \log C(r)}{\partial \log r}$, which implies that the correlation dimension is $n(r)$ at $r=0$. In this sense, $n(r)$ extends the correlation dimension from only being defined at scale $r=0$ to positive scales. 
\end{remark}

\begin{remark}
The assumption that $n_m(r)$, $n_s(r)$ are slowly varying functions of $r$ ensures that they are almost constants in small intervals and therefore can be estimated by counting points in these intervals (see Sect. \ref{sec:5}). 
\end{remark}

\subsection{Preserving the growth rate of the relative capacity}
With the rigorous definition of capacity, let us explain the crowding issue using the manifold $\mathcal{M} = B^n_2(1)$ endowed with the uniform distribution. Again, consider the ball $B^n_2(1)$ as a union of concentric shells with infinitesimal thickness and increasing radii.  Then the relative density of the shell with radius $r$ is $\rho( r; \mathcal{M},f_{uniform}, \|\cdot \|_2) \sim nr^{n-1}$. If these shells are mapped to $d=2, 3$ without changing their inclusion order, then the relative density in the low dimension (say $d=2$) is $\rho(r; \mathcal{S}, f_{uniform}, \|\cdot\|_2)  \sim 2r^{1}$. The crowding issue arises because the two relative densities grow with $r$ at different rates, and more specifically, because the relative density in the original (high) dimension grows faster with $r$ than that of the low dimensional one,  points will be crowded near the center.

%

 To solve this problem, we need to make these two rates match by either 1) equipping the low dimensional manifold with non-uniform probability distributions so as to accelerate its capacity growth rate, or 2) replacing the metric in the high dimensional space with another metric so as to slow down its capacity growth rate.  We go with the latter option in this paper since the resulting visualization is more digestible. 

Explicitly, we want to design a new distance $\hat{D}$ such that  the relative densities of the original manifold and the embedded one can approximately match 
\[
\rho(r; \mathcal{M},f, \hat D) \approx  \rho(r; \mathcal{S}, f_{uniform}, \|\cdot\|_2) 
\]
for all $r\geq 0$. Since $\rho(r)$ is the derivative of $C(r)$,  this in turn implies
\[
C(r; \mathcal{M},f, \hat D) \approx C(r; \mathcal{S}, f_{uniform}, \|\cdot\|_2).
\]
The following theorem provides a way to define the new distance that ensures the match.
\begin{theorem}\label{thm:dist} Let $\mathcal{M}$ be the high dimensional data manifold and $\mathcal{S}$ be the embedded one. Let $P: \mathcal{M} \rightarrow \mathcal{S}$ be the dimension reduction mapping. For any pair of data point $(x,z)$,  suppose  their distance after embedding is an (arbitrary) function of their distance before embedding, i.e., $\| P(x)-P(z)\|_2 = G(\|x-z\|)$ where  $G$ is some unknown function and $||\cdot\|$ is the original metric equipped to $\mathcal{M}$. Under Assumption 1 and Assumption 2,  we can define a new ``distance'' between any pair of points $x$ and $z$ as
\begin{equation}\label{eq:dist}
\hat{D}(\|x-z\|) :=  \|x-z\|^{\frac{n_m (\|x-z\|)}{n_s(\|P(x)-P(z))\|_2)}},  \quad x, z \in \mathbb{R}^n.
\end{equation}
where $n_m(\cdot)$ and $n_s(\cdot)$ are the same as defined in Assumption 2. 
This distance allows a match in the relative capacities between the original manifold and the embedded manifold endowed with the uniform density and the usual $\ell_2$ metric, i.e.,
\[
C(r; \mathcal{M},f, \hat{D})=C(r; \mathcal{S},f_{uniform}, \|\cdot \|_2), 
\]
The matching of the capacities means that for a certain region, its density in the visualization reflects its true density in the original dataset, hence no crowding would occur.
\end{theorem}
The proof can be found in the appendix.
\begin{remark}\label{rm:1} When the distances are measured under $\hat{D}$ in the high dimensional space, this theorem ensures that points would neither collapse nor be pulled apart during embedding. Then the question is how well this new ``distance'' $\hat{D}$ can represent the original distance $\|\cdot \|$ equipped to the manifold? For the preservation of geometry, we hope $\hat{D}$ to be a monotonically increasing function of the original metric, so that larger distances are still larger and smaller distances still smaller.
Unfortunately, one immediate sees that this monotonicity cannot be guaranteed by the current definition of $\hat{D}$. Since by the proof of Theorem \ref{thm:dist}, the current definition of $\hat{D}$ is both necessary and sufficient for a match of the relative capacity, we know that simultaneously preserving the capacity and the monotonicity is impossible. Therefore, in practice, we propose to  bring back the monotonicity  with a small change to the current $\hat{D}$. We will define the modified distance $\widetilde{D}$ from the following recursive procedure. 1). order all the pairwise distances computed from the original metric on $\mathcal{M}$. 2). For the smallest pairwise distance, say $\sigma_{0}$, simply set $\widetilde{D}(\sigma_{0}) = \hat{D}(\sigma_{0})$.  For the second smallest distance, say  $\sigma_{1}$, we need to define $\widetilde{D}(\sigma_1)$ as close to  $\hat{D}(\sigma_1)$ as possible while having the monotonicity $\widetilde{D}(\sigma_1)\geq \tilde{D}(\sigma_0)$. Thus we set $\widetilde{D}(\sigma_1)=\max\{\hat{D}(\sigma_1), \widetilde{D}(\sigma_0)\}$. Assume we have defined $\widetilde{D}(\sigma_k)$ that has monotonicity up to $\sigma_k$,  setting $\widetilde{D}(\sigma_{k+1})$ by
\begin{equation}\label{eq:new_dist}\widetilde D(\sigma_{k+1}) = \max\{\hat{D}(\sigma_{k+1}), \widetilde{D}(\sigma_k)\} \end{equation}
guarantees $\widetilde{D}(\sigma_{k+1})$ to be monotonic up to $\sigma_{k+1}$.



Intuitively, the maximum makes $\widetilde D$ assign more room to the low dimensional space than necessary. In practice, assigning excessive room is much less harmful than not assigning enough room, because the latter leads to the crowding issue in the visualization.
\end{remark}
\begin{definition} We call the new ``distance'' defined in \eqref{eq:dist} as the Capacity Adjusted Distance (CAD), and the modified version $\tilde{D}$ defined in \eqref{eq:new_dist} as the modified Capacity Adjusted Distance (modified CAD).
\end{definition}
\begin{remark}\label{rm:dist} In Theorem \ref{eq:dist},  the original distance metric $\|\cdot\|$ associated with the high dimensional manifold $\mathcal{M}$ could be Euclidean distance,  geodesic distance, diffusion distance or others.  We make the choice of distances an option to the user in our algorithm (Algorithm \ref{alg:1}). \end{remark}

The definition of $\widetilde{D}$ requires the knowledge of the intrinsic dimension $n_m(r)$ and $n_s(r)$  at various scales $r$. In the next section, we introduce a dimension estimation method that allows an estimate of $n_m(r)$ from the dataset $\{X_i\}_{i=1}^N$. For $n_s(r)$, since we used uniform distribution along with the $\ell_2$ metric on the embedded manifold,  in light of Example 1, this means $C(r, \mathcal{S}, f_I, \|\cdot\|_2) \simeq r^{d}$, hence $n_s(r)=d$. 

\subsection{Estimation of $n_m(r)$ at various scales}\label{sec:5}
  We did not find an existing method that can calculate the multi-scale correlation dimension for all scales. The only similar work that we are aware of is \cite{lee2015multi}, which calculates the \textit{average} dimension up to scale $r$, while here we need the instantaneous dimension at scale $r$.  We propose the following way to estimate the dimension function $n(r)$ from the data. Recall that in Assumption 2, $n(r)$ is defined as the dimension at scale $r$ and is playing a role in the definition of the relative density  $\rho(r)$ (Assumption 2):
 \begin{equation}\label{eq:ball} \rho(r) \equiv \rho(r; \mathcal{M},f, \|\cdot \|_2) = cn(r)r^{n(r)-1} \end{equation}
 where $c$ is some unknown absolute constant and $n(r)$ is a slowly varying with respect to $r$. On a small interval $[r_0, r_0+\Delta r]$, we can assume $n(r)$ to be a constant, denoted by $n\equiv n(r_0)$. Then $\rho(r)$ becomes 
   \begin{equation}\label{eq:ball} \rho(r) \equiv \rho(r; \mathcal{M},f, \|\cdot \|_2) = cnr^{n-1},  \quad  r\in [r_0, r_0+\Delta r]. \end{equation}
  We can estimate $\rho(r)$ by taking finite difference of $C(r)$ with respect to $r$ while the latter can be estimated from the data using the counting number. Specifically, by definition, the normalized counting number $\hat{C}_N(r)$ below is a consistent estimate of $C(r)$, i.e., $\hat{C}_N(r) \rightarrow C(r)$ as $N\rightarrow \infty$. 
    \begin{equation}\label{eq:hatC}
 \hat{C}_N(r) :=   \frac{1}{N} \cdot \frac{1}{N-1} \sum_{i=1}^N\sum_{j=1,j\neq i}^N \chi_{\{\hat D(X_i, X_j) \leq r \}}.
\end{equation}
With  $\hat{C}_N(r) $ computed from data, we can estimate $\rho(r)$ by the finite difference
\[
\hat{\rho}_N(r) = \frac{\hat{C}_N(r+\Delta r)-\hat{C}_N(r)}{\Delta r}.
\]
When $\hat{\rho}_N(r)$ is known on $[r_0, r_0+\Delta r]$, we can then compute $n$ by fitting the slope of $\log \hat \rho$
\begin{equation}\label{eq:initial}
n \approx \frac{\log(\hat \rho_N(r_0+\Delta r))-\log(\hat \rho_N(r_0))}{\log (r_0+\Delta r)-\log(r_0)}.
\end{equation}

However, this estimate is unstable especially when the data is insufficient and may even produce a negative dimension. Therefore, we only use this estimate as an initial guess for solving $n$ from \eqref{eq:ball}. In order to use \eqref{eq:ball}, we first need to estimate $c$ by looking at the equation at $r=0$. As mentioned in Remark \ref{rm:corr}, $n(r)\vert_{r=0}$ corresponds to the well-known correlation dimension, and can be estimated by counting points in the ball $B(0,\Delta r)$ with decreasing radius $\Delta r \rightarrow 0$ and then be solved from 
\begin{equation}\label{Eq:c} C(\Delta r) =c\Delta r^{n(0)}
\end{equation}
 via a linear fitting procedure between $\log C$ and $\log \Delta r$  [3, 5, 12], where $C(r)$ can again be approximated by the counting number $\hat{C}_N(r)$. Once $n(0)$ is computed, plugging it into \eqref{Eq:c} to get $c$. With the estimated value $\hat{c}$, we can then numerically search for the $n$  that makes \eqref{eq:ball} best satisfied, that is, we solve the following optimization problem for $n$
 \[
 \hat{n}(r_0) = \arg\min\limits_n \|\hat{\rho}_N(r_0) - \hat{c} nr_0^n \|_2^2
 \]
 since this optimization problem only has one degree of freedom, we can perform brute force search for the dimension.
\subsection{Dimensionality reduction with the modified Capacity Adjusted Distance}\label{sec:4.2}
With the well-defined adjusted-CAD distance $\widetilde{D}$, we propose the following dimensionality reduction procedure.  We search for the low dimensional vectors $\{Y_i\}_{i=1}^N$ that  best preserves $\widetilde{D}$ under the Kullback–Leibler divergence, 
\begin{equation}\label{eq:KL} \{\hat Y_i\}_{i=1}^N =\arg\min\limits_{Y_i,i=1...,N} \sum_{i,j} p_{i,j} \log \frac{p_{i,j}}{q_{i,j}}  \end{equation}
where
\begin{equation}\label{eq:prob1}
p_{i,j} = \frac{ (\epsilon+\widetilde{D}^2(\|X_i-X_j\|))^{-1}}{\sum_{k,l, k\neq l} (\epsilon+\widetilde{D}^2(\|X_k-X_l\|))^{-1}},  \quad q_{i,j} = \frac{(1+\|Y_i-Y_j\|^2)^{-1}}{\sum_{k,l k\neq l} (1+\|Y_k-Y_l\|^2)^{-1}},
\end{equation}
where the reciprocal and the normalization together transform the distances into probabilities, $\|\cdot\|$ is the original distance metric in the high dimensional space, and $\epsilon>0$ is some small constant used to avoid taking the reciprocal of 0.  Putting the $p_{i,j}$ defined in \eqref{eq:prob1} into a matrix,  $P=[p_{i,j}]_{i,j=1}^N$. We can think of this $P$ as the normalized probability matrix associated with a random graph constructed as follows. Let the $N$ data points $\{X_i\}_{i=1}^N$ correspond to the $N$ nodes of the graph, and  node $i$ is connected with node $j$ by an edge with probability $p_{i,j} \sim \widetilde{D}^{-2}(X_i,X_j)$, where $\sim$ hides a universal constant (i.e., the denominator in \eqref{eq:prob1}). This is to say, closer  points are more likely to be connected by an edge in this random graph. The optimization \eqref{eq:KL} is hence trying to match the KL divergence between the probabilities of the original and the embedded graphs. 

The formulation \eqref{eq:KL} looks similar to that of t-SNE
\begin{equation}\label{eq:tSNE} \textbf{t-SNE:} \quad \{\hat Y_i\}_{i=1}^N =\arg\min\limits_{Y_i,i=1...,N} \sum_{i,j} p_{i|j} \log \frac{p_{i|j}}{q_{i|j}}  =\arg\min\limits_{Y_i,i=1...,N} \sum_{i,j} - p_{i|j} \log q_{i|j}\end{equation}
where
\begin{equation}\label{eq:pro}
p_{i|j} = \frac{ e^{- \frac{\|X_i-X_j\|_2^2}{2\sigma^2 }}}{{\sum_{k, k\neq j}} e^{- \frac{\|X_k-X_j\|_2^2}{2\sigma^2}}},  \quad q_{i|j} =  \frac{(1+\|Y_i-Y_j\|_2^2)^{-1}}{\sum_{k, k\neq j} (1+\|Y_k-Y_j\|_2^2)^{-1}},
\end{equation}
Compared to t-SNE, our formulation \eqref{eq:KL} has the following merits.
\begin{enumerate}
\item \textbf{Less over-streching}: t-SNE mitigates the crowding issue and promotes the formation of clusters by matching Gaussian distributions with t-distributions. Intuitively, this makes close points closer and far away points  further, but the degree of stretching is had to control. As a result, t-SNE may produce fake clusters due to the over-stretching.   In contrast, our method performs the right amount of stretching that is necessary to avoid the crowding. In other words, the stretching in our method is much milder. This can also be seen from the definition of $p_{i,j}$ and $q_{i,j}$ in \eqref{eq:prob1}, which are both heavy tailed distributions (recall that the adjusted distance $\widetilde{D}(\|X_i-X_j\|)$ has the form of  $\|X_i-X_j\|^{\alpha}$ with some $\alpha$ so $p_{i,j}$ is also heavy tailed), whereas in t-SNE, the $t$-distribution is fitting with the light tailed Gaussian distribution, therefore creating more stretching.  
\item \textbf{No tuning parameter:} the performance of t-SNE heavily depends on the choice of the bandwidth parameter $\sigma$, whereas our method does not have a key tuning parameter (the small positive constant $\epsilon$ used to avoid dividing by 0 does not affect the results much as long as it is sufficiently small). 
\item\textbf{Better at preserving the geometry:} In the objective function of t-SNE, a conditional probability $p_{i|j}$ is used,  i.e., $\sum_{i} p_{i|j} =1, \forall j$, which means for each $j$, the sum of the similarities between $X_j$ and all other points is normalized to 1. Therefore, t-SNE will produce an embedding in which all points are about equally far from the entire dataset. If the original dataset does not have this property, then t-SNE will distort its geometry.  For example, for a dataset with outliers, the outliers would be further away from the entire dataset than points at the center, so the aforementioned equal distance property is violated. But t-SNE will nonetheless impose this property in its embedding, hence after embedding one can no longer tell who are the outliers and who are the center points.
For datasets with cluster structures, this further means after t-SNE embedding, clusters of different sizes  become of similar sizes  and those with different distances now have similar distances \footnote{We refer the readers to the website https://distill.pub/2016/misread-tsne/ for more such examples.}. The cluster size and distance information is lost. The same happens for SNE and UMAP. In contrast, the probabilities in our formulation are not conditional probabilities, therefore will not suffer from this type of distortion. Mathematically speaking, we do not normalize each row of the probability matrix $P$ individually, but normalizing the entire matrix $P$ by one constant. The variation among rows survives and carries the correct geometric information.  It is worth noting that this universal normalization does not apply to t-SNE ( or SNE or UMAP) because of their usage of the Gaussian kernel. The fast decay of the Gaussian tail often causes the sum of certain rows of $P$ to be way smaller than others. As the row sums are the weights in front of the unknown variables $Y_i$ in the objective function, if they are too small, the gradient descent algorithm will not update the corresponding $Y_i$ much hence causing a very slow convergence or even a wrong solution.

\end{enumerate}
%

\begin{algorithm}[H]
\SetAlgoLined
\KwOut{low dimension embedding $Y_i$, $i=1,...,N$ }
\KwIn{original data $X_i$, $i=1,...,N$; target dimension $d = 2$ or 3; choices of the high dimensional distance $\|\cdot\|$: Euclidean, geodesic or diffusion; number of scales: $M$}
 \For{r = 1: M}{
  estimate the dimension $n_m(r)$ from $\{X_i\}_{i=1}^N$ via solving \eqref{eq:final} \;}
  \For{i=1:N}{
  \For {j=i+1:N}{
   
   use the estimated $n_m(r)$ at $r=\|X_i-X_j\|$  to defined the modified distance $\widetilde{D}(\|X_i- X_j\|)$ as in \eqref{eq:new_dist}  \;
   }
   }
  Solve the optimization \eqref{eq:KL} to construct $\{Y_i\}_{i=1}^N$.
 \caption{Capacity Preserving Mapping (CPM)}
 \label{alg:1}
\end{algorithm}
\subsection{Comparisons to other methods}\label{sec:comparison}
Besides t-SNE,  our method is also related to the non-metric MDS (NMDS) \cite{rabinowitz1975introduction} and the multi-scale SNE \cite{lee2015multi}.
Similar to our method, the non-metric MDS also aims at preserving the pairwise dissimilarity as  closely as possible. It approaches this goal by minimizing a scaled distances between points. Let $d_{i,j}$ be the high dimensional distances, the NMDS embedding $\{Y_i\}_{i=1}^N$ is obtained by solving the optimization problem 
\begin{equation}\label{eq:nmds}
\min_{\{Y_i\}_{i=1}^N, f\in \mathcal{F}} \sum_{i,j}  | \|Y_i-Y_j\|_2- f(d_{i,j})|
\end{equation}
where $\mathcal{F}$ is the set of positive monotonically increasing functions. The scaling function $f$ plays the role of mitigating the crowding. Indeed, the scaled distance $f(d_{i,j})$ essentially corresponds to our capacity adjusted distance $\widetilde{D}(d_{i,j})$. From this perspective, our formulation provides an explicit way to compute $f$ which avoids the trouble of solving it from an optimization. In addition, the NMDS uses the $\ell_1$ norm instead of the KL divergence in the objective, so it fails to render the correct small-scale information of data (see Figure \ref{figure_3}).  Replacing the $\ell_1$ norm in \eqref{eq:nmds} with a KL-divergence type of dissimilarity measure is also not plausible as it will make the optimization too difficult to solve.

Multi-scale SNE \cite{lee2015multi} is similar to our approach in the sense that both methods assume a scale-varying dimension of the data manifold. However, since it inherits the structure of SNE, the third drawback  mentioned in the previous subsection applies. 

\section{Numerical simulation}
By construction, the proposed Capacity Preserving Mapping (CPM) algorithm does not promote formation of clusters, so it will not produce as well separated clusters as tSNE and UMAP.  But our method respects the geometry of the original dataset and therefore would be a good supplement to the various clustering algorithms.

To demonstrate how CPM performs and how to read the results, we first compare it with the landmark methods non-metric MDS, Isomap, and t-SNE (since t-SNE and UMAP produce quite similar results for all these experiments in the sense that they are about equally good in clustering and not so good at preserving the geometry, here we only show the results of t-SNE). We consider four datasets: 1) the motivating example introduced in Sect. \ref{sec:3}, 2) the augmented Swiss roll to be defined shortly, and 3) the MNIST dataset 4) Coil 20.
%

\begin{figure}[htbp]
\includegraphics[height=10.0cm,width=16.5cm]{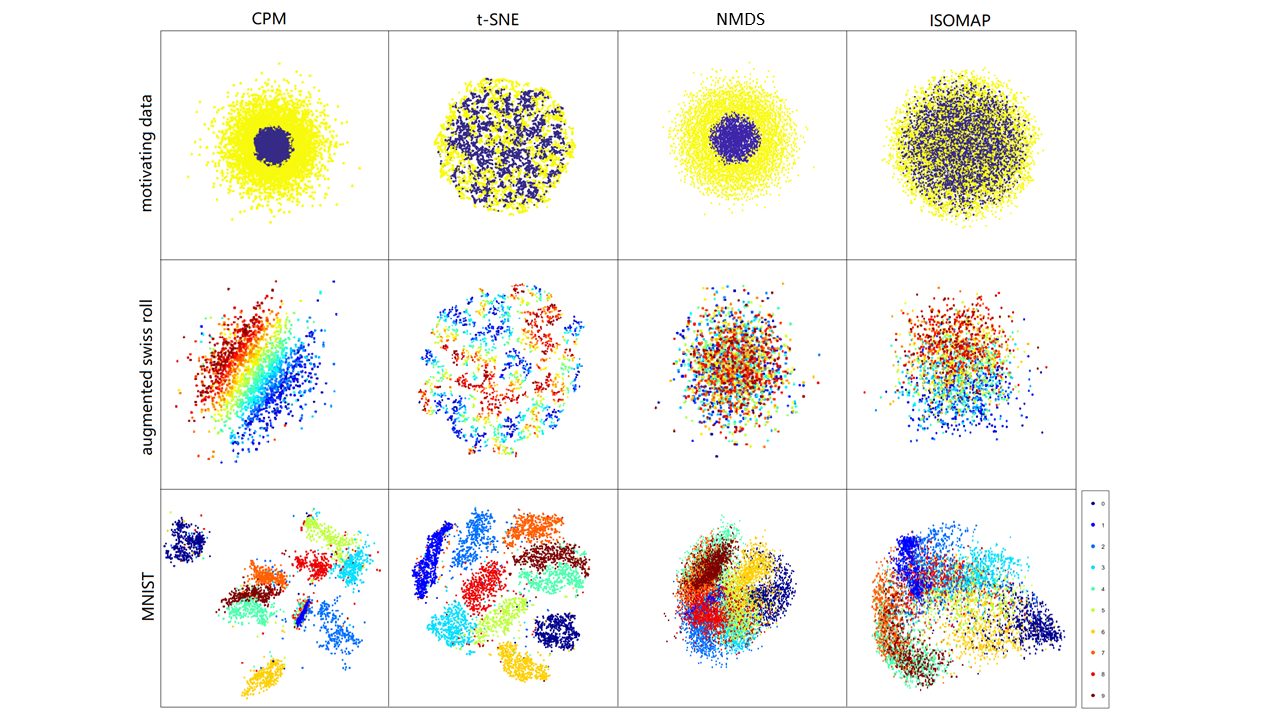}
\caption{A comparision of CPM, t-SNE, NMDS and ISOMAP}
\label{figure2a}
\end{figure}
\begin{figure}[htbp]
\centering
\includegraphics[scale=0.5]{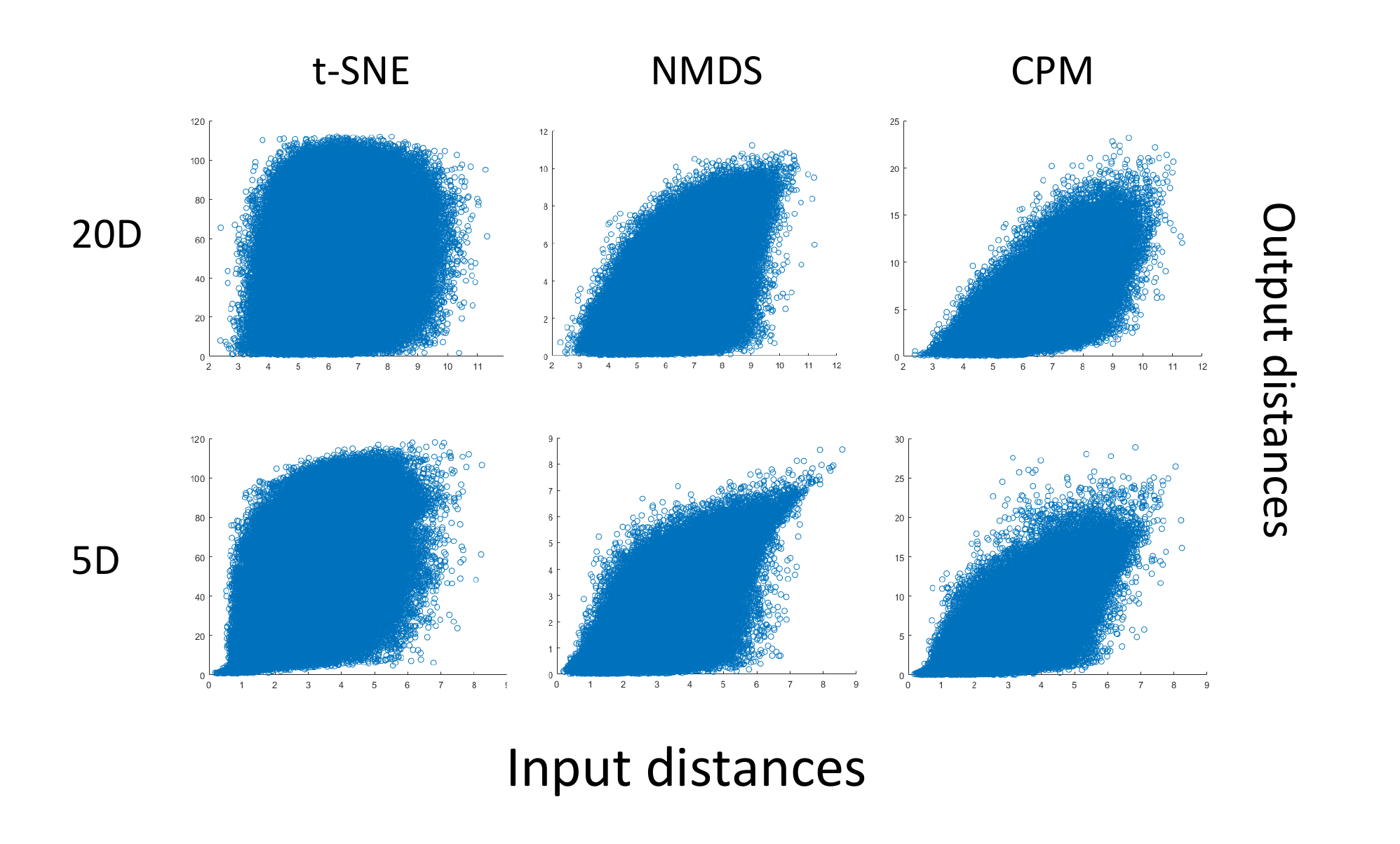}
\caption{Shepard diagrams of t-SNE, NMDS and CPM applied to Gaussian point clouds}
\label{figure_3}
\end{figure}

\textbf{The motivating example:}
in Sect. \ref{sec:3}, we observed the crowding phenomenon when using MDS to map two non-overlapping objects in $\mathbb{R}^5$ to $\mathbb{R}^2$, with one object being an $\ell_2$ ball and the other being a shell lying right outside the ball. 
The first row of Figure \ref{figure2a} shows the mapping results of this motiving example by four other methods. We can see that the crowding problem arises in t-SNE (perplexity =$30$), and Isomap.  In contrast, CPM and non-metric MDS (NMDS) are able to mitigate the crowding and reveal the correct relation between the two classes. To further compare CPM and NMDS, we plot the Shepard diagram, which shows the goodness-of-fit by plotting the allpair distances before versus after embedding. Ideally, the ranking of these distances should be preserved and one observes a monotonically increasing curve in Shepard diagram. However, in practice, one usually sees a ``thick'' curve due to the distortion.  Figure \ref{figure_3} displays the Shepard diagrams of NMDS, CPM and  t-SNE (perplexity 30, other perplexity values produce similar or worse results) when mapping 5 dimensional (bottom row) and 20 dimensional (top row) Gaussian point clouds (each containing 1000 i.i.d. sampled points according to $N(0, I_d)$, with $d=5,20$, respectively) down to 2D. We see that t-SNE is only good at preserving small scale distances. NMDS is good at preserving medium and large distances, but is not as good as t-SNE in preserving the small scale ones. CPM has a similar performance to t-SNE at small scales and a similar performance to NMDS at medium and large scales and the advantage of CPM becomes more visible as the dimension gets higher (i.e., in the 20 dimensional case).   


\textbf{Augmented Swiss roll:}
the Swiss roll is a popular synthetic test dataset that can evaluate a method's ability in preserving the geometric structure. However, Swiss roll is a 2 dimensional dataset,  so it cannot test the ability of an algorithm in mapping the data to below its intrinsic dimension. Therefore,  we construct the augmented Swiss roll dataset $X=[x_1,...,x_p]$, where the first three coordinates $[x_1, x_2, x_3]$ are  the original Swiss roll, and the rest of the coordinates $x_4...,x_p$ are filled with i.i.d. Gaussian entries.  Therefore the intrinsic dimension of $X$ is $p-1$. Here we set $p=6$. Explicitly, we set
\begin{align*}
x_1(t) & = (t+1)\cos(t), \\
x_2(t) & = (t+1)\sin(t), \\
x_j(t) & = g_j(t), \ \ j=3,...,6
\end{align*}
where $t$ takes random values on the interval $[0,1]$, and for $j=3,...,p$ and any $t$, $g_j(t)$ are i.i.d. Gaussian variables with mean 0 and variance 25.

To map this dataset to 2D for visualization, we apply Isomap, NMDS with geodesic distance, and CPM with geodesic distance, all with the same number of neighbours (=10). From the second row of Figure \ref{figure2a}, we see that in terms of unfolding the manifold, CPM did the best job among all. The result of t-SNE (perplexity 30) is also included for comparison.
%

\textbf{MNIST dataset} We repeat the previous experiment on real data. The MNIST dataset contains  60000 training images of handwritten digits. We apply t-SNE, Isomap, NMDS with geodesic distance and CPM with geodesic distance to a subset of 6000 randomly selected images from the training set. As mentioned before, we do not expect CPM to produce as nice-looking clusters as t-SNE because it is not designed as a clustering algorithm. Nevertheless, the separation of clusters in CPM is pretty good, see the bottom row of Figure \ref{figure2a}. Since CPM is not stretching the data, this nice separation tells us that the original dataset already has pretty well separated clusters. In addition, the CPM result reveals some new information about the dataset. For example, it shows that the cluster of digit 1 has the smallest variance, which is consistent with our intuition that the handwritten digit 1 has the least variation among different writing styles. To confirm this observation, we computed the variances of the clusters based on their true high-dimensional representation and put the results in Table \ref{figure2f}. We can see that the cluster of digit 1 indeed has a much smaller variance than all other clusters. The second and third smallest clusters are digit 7 and digit 9, which also appear to be smaller than others in the visualization.  Another piece of information conveyed by the CPM visualization is that the cluster of digit 1 is close to many other clusters (so it is at the center), which is aligned with the intuition that the handwritten digit 1 looks similar to 2, 3, 7, and maybe 9. All these pieces of information are lost in the t-SNE.
 
\begin{table}[htbp]
\caption{Variances of clusters in MNIST dataset (normalized)}
\label{figure2f}
\renewcommand\arraystretch{2}
\begin{tabular}[c]{|c|c|c|c|c|c|c|c|c|c|c|}
\hline
Digit & 0 & 1 & 2 & 3 & 4 & 5 & 6 & 7 & 8 & 9 \\
\hline
Variance & 0.991 & 0.448 & 1.000 & 0.880 & 0.824 & 0.952 & 0.871 & 0.750 & 0.891 & 0.756\\
\hline
\end{tabular}
\end{table}

\subsection{Coil 20} 
The Coil 20 dataset contains images of 20 objects captured from different angles while they rotate.  Previous methods are only aiming at separating the images into 20 clusters.  Here we also care about the shape of clusters. In particular, we check if the following properties are preserved in the embedding,
\begin{enumerate}
\item of the 20 objects, those that look more different from different angles should correspond to clusters with large sizes in the visualization; similarly, if an object is nearly isotropic (looks similar from all angles), then its corresponding cluster should have a very small size. 
\item objects similar to each other should correspond to clusters close to each other in the visualization;
\item if an object is symmetric with respect to its center, then the corresponding data should form a trajectory similar to a folded circle in the visualization;

\end{enumerate}
We now evaluate the performance of CPM based on these three criteria. 

\begin{figure}[htbp]
\centering
\includegraphics[height=7.0cm,width=10cm]{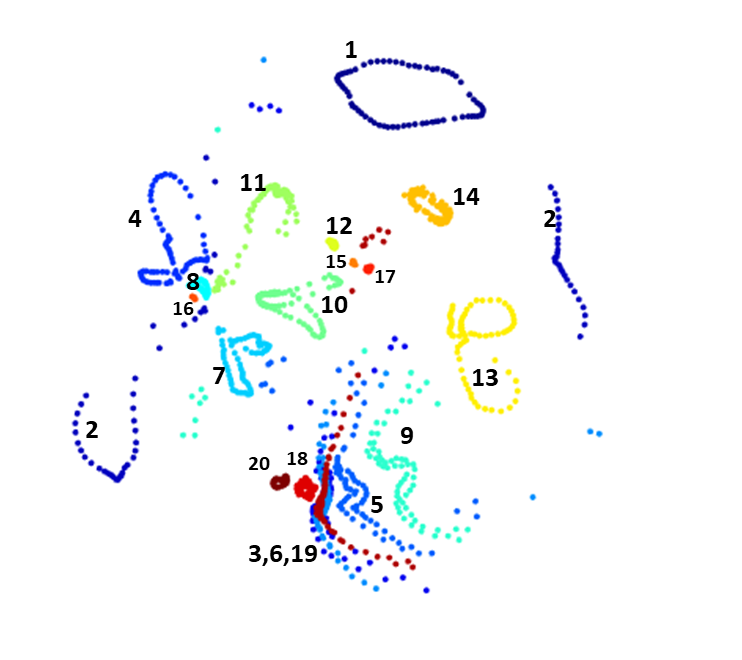}
\caption{Visualization by CPM.}
\label{fig:figure2b}
\end{figure}

1. Table \ref{table:3} summarizes the variance of each object as it rotates. The variance $V_i$ for the $i$th object is computed using the formula
\[
V_i =  \frac{1}{n_i} \sum_{j=1}^{n_i} \|X_j - \bar X_i\|_2^2
\]
where $n_i$ is the number of points in the $i$th cluster,  $X_j$, $j=1,...,n_i$ are vectorized images of the $i$th object, and $\bar{X}_i=\frac{1}{n_i}\sum\limits_{j=1}^{n_i} X_j$ is the mean.  Table \ref{table:4} and Figure \ref{fig:figure2b} together confirm that objects with large variances during rotations (i.e., less isotropic) corresponding to clusters with large sizes in the visualization.

\begin{center}
\begin{figure}[htbp]
\centering
\includegraphics[height=6.0cm,width=7cm]{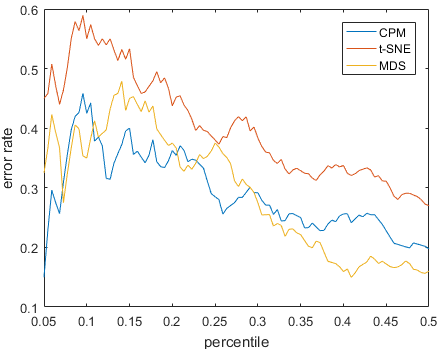}
\caption{Error rates in preserving relative cluster proximity by CPM,t-SNE and MDS}
\label{fig:dissim} 
\end{figure}
\end{center}

2. To characterize the preservation of the inter-cluster distances of the 20 classes/objects in Coil 20,  we propose to use the distance error index defined as follows.   For a given cluster say Cluster $i$, rank the other 19 clusters according to their distances to Cluster $i$ in increasing order (the distance between two clusters are defined as the averaged pairwise distances between one cluster and the other). For a fixed value $p$, the clusters in the first $p$th percentile of this ranking are called a neighbour to Cluster $i$, otherwise is a non-neighbour of Cluster $i$.  After doing this for each $i=1,...,20$,  we build a $20\times 20$ proximity matrix $C(p)$, where $C_{i,j}(p) =1$ if Cluster $j$ is a neighbour of Cluster $i$ (within the $p$th percentile), and $C_{i,j}(p) =0$ otherwise.  We compute the proximity matrix of the original data (obtain $C^{orig}$) and that of the embedded data (obtain $C^{emb}$), and then compute the dissimilarity between $C^{orig}$ and $C^{emb}$:
\[
error(p) = \frac{\sum_{i,j} 1_{\{C^{orig}_{i,j}(p)=1, C^{emb}_{i,j}(p) =0\}}}{\sum_{i,j} 1_{\{C_{i,j}^{orig}(p)=1} \}}
\]  
That is to say, $error(p)$ contains the percentage of neighbouring clusters that are no longer neighbours the embedding. For any predefined percentile $p$,  we can obtain a value for $error(p)$. Let  $p$ range from 0\%-50\%,  we derive the percentile versus dissimilarity plot in Figure \ref{fig:dissim} for three embedding methods: CPM with Euclidean distance, t-SNE (perplexity 30) and MDS. Clearly, MDS is good at preserving very large inter-cluster distances,  our method is better at preserving small to middle scale inter-cluster distances, and t-SNE is not good at preserving inter-cluster distances.

3. Table \ref{table:6} shows that the CPM embedding indeed produces folded circles for symmetric objects.

\begin{table}[htbp]
\caption{Variances of clusters in the 2D visualization of the COIL 20 dataset}
\renewcommand\arraystretch{2}
\begin{tabular}[c]{|c|c|c|c|c|c|c|c|c|c|c|}
\hline
Label & 1 & 2 & 3 & 4 & 5 & 6 & 7 & 8 & 9 & 10\\
\hline
Variance & 51.19 & 69.92 & 38.42 & 32.83 & 42.55 & 48.58 & 25.39 & 12.34 & 45.69 & 27.34\\
\hline
Label & 11 & 12 & 13 & 14 & 15 & 16 & 17 & 18 & 19 & 20\\
\hline
Variance & 28.31 & 3.45 & 31.72 & 21.10 & 2.03 & 1.90 & 2.97 & 10.72 & 43.18 & 8.37 \\
\hline
\end{tabular}
\label{table:3}
\end{table}

\begin{minipage}{\textwidth}
 \begin{minipage}[t]{0.5\textwidth}
  \centering
   \makeatletter\def\@captype{table}\makeatother \caption{Objects with large variances}
	\renewcommand\arraystretch{2}
	\begin{tabular}[c]{|c|c|c|c|c|}
	\hline
	Objects & {\includegraphics[scale=0.20]{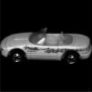}} & {\includegraphics[scale=0.20]{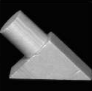}} & {\includegraphics[scale=0.20]{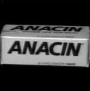}} & {\includegraphics[scale=0.20]{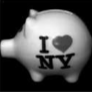}}\\
	\hline
	Labels & 19 & 2 & 5 & 13 \\
	\hline
	Variance & 43.18 & 69.92 & 42.55 & 31.72\\
	\hline
	\end{tabular}
	 \label{table:4}
 \end{minipage}
 \begin{minipage}[t]{0.5\textwidth}
  \centering
   \makeatletter\def\@captype{table}\makeatother \caption{Objects with small variances}
	\renewcommand\arraystretch{2}
	\begin{tabular}[c]{|c|c|c|c|c|}
	\hline
	Objects & {\includegraphics[scale=0.20]{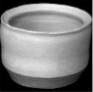}} & {\includegraphics[scale=0.20]{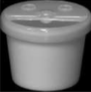}} & {\includegraphics[scale=0.20]{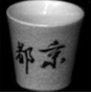}} & {\includegraphics[scale=0.20]{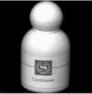}}\\
	\hline
	Labels & 17 & 15 & 12 & 16 \\
	\hline
	Variance & 2.97 & 2.03 & 3.45 & 1.90\\
	\hline
	\end{tabular}
	 \label{table:5}
 \end{minipage}
\end{minipage}
\begin{table}[htbp]
\centering
\caption{Visualizations of symmetric objects}
\renewcommand\arraystretch{1.5}
\begin{tabular}[b]{|c|c|c|c|c|c|}
\hline
Objects & {\includegraphics[scale=0.3]{5.png}} & {\includegraphics[scale=0.3]{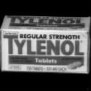}} & {\includegraphics[scale=0.3]{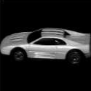}} & {\includegraphics[scale=0.3]{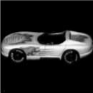}} & {\includegraphics[scale=0.3]{19.png}}\\
\hline
Visualizations & {\includegraphics[scale=0.2]{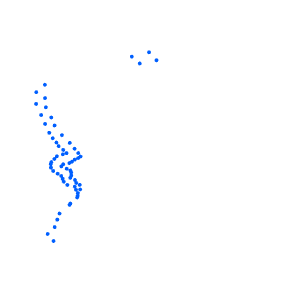}} & {\includegraphics[scale=0.2]{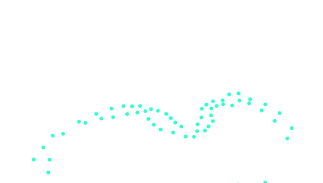}} & {\includegraphics[scale=0.2]{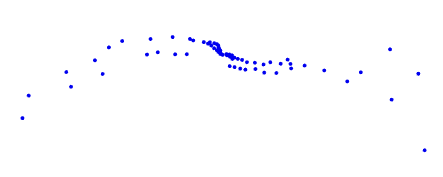}} & {\includegraphics[scale=0.18]{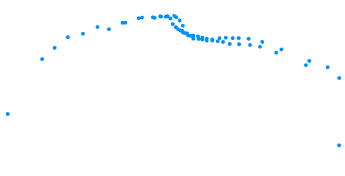}} & {\includegraphics[scale=0.12]{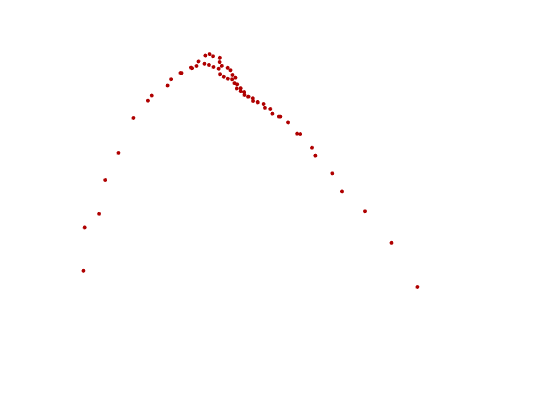}}\\
\hline
\end{tabular}
\label{table:6}
\end{table}

\section{Conclusion and future directions}
In this paper, we answered the question of how to rigorously characterize and treat the intrinsic crowding issue in data visualization. After giving a mathematical notation to the capacity, we discussed two possible directions to mitigate the crowding: altering the density or altering the distance. We chose the latter for simplicity, but it will be interesting to explore the former as well. 

After the Capacity Adjusted Distance was defined, we proposed to find the low dimensional embedding by matching the dissimilarity measured by the KL divergence. There are many other dissimilarity measures in the literature. The performance of the combination of the Capacity Adjusted Distance with other measures is yet to be explored. 


In the numerical experiments, we only tested CPM under Euclidean and Geodesic distances. It would be interesting to see how it works with other distances (e.g., the diffusion distance).

\section{Appendix}
\subsection{Proof of Theorem \ref{thm:dist}  }
\begin{proof}
The assumption that the embedded pairwise distances can be written as a function of the pairwise distances in the original space, i.e., $\|P(x)-P(z)\|_2 = G(\|x-z\|)$ (with some function $G$), allows  us to define the new distance $\hat{D}(x,z)$ to be a function of $\|x-z\|_2$ only, that is $\hat{D}(x,z)$ can be written as $\hat{D}(||x-z||)$ for simplicity. We will show that if $\hat{D}$ is defined as in the theorem, i.e., $\hat{D}(\|x-z\|) = {\|x-z\|}^{\frac{n_m(\|x-z\|)}{n_s(\|P(x)-P(z)\|_2)}}$, then the capacity is preserved. Without any ambiguity from the context, we write $n_m(\|x-z\|)$, $n_s(\|P(x)-P(z)\|_2)$ in short as $n_m$ and $n_s$. By the definition of the relative capacity, for any $\tilde{r}>0$, and infinitesimal $d\tilde r$, we have
\begin{align*}
C(\tilde{r}+d \tilde{r}; \mathcal{M},f, \hat{D})-C(\tilde{r}; \mathcal{M},f, \hat{D}) &= \mathbb{P}_{x,z\sim f(\mathcal{M})} (\tilde{r}\leq \hat{D}(\|x-z\|_2)\leq \tilde{r}+d \tilde{r})\\
&=p_{x,z\sim f(\mathcal{M})} (\hat{D}(\|x-z\|_2)= \tilde{r})d \tilde{r}\\
& =\rho(\tilde{r}; \mathcal{M},f, \hat{D})d \tilde{r} \\
\end{align*}
where $p_{x,z}$ denotes the probability density function with respect to the random variables $x$ and $z$. The first and last equalities used the definitions of $C(r;\mathcal{M},f,\hat{D})$ and $\rho(\tilde{r}; \mathcal{M},f, \hat{D})$. Hence the relative capacities match if the relative densities match. We will show that relative densities match by showing 
\begin{equation}\label{eq:toshow}\rho(\tilde{r}; \mathcal{M},f, \hat{D})d \tilde{r} =c n_s \tilde{r}^{n_s-1} d\tilde{r},\end{equation} because the right hand side is exactly the relative density for the low dimensional embedding $\rho( \tilde{r}; \mathcal{S},f_I, \|\cdot\|_2)$ by Assumption 2. To compute the left hand side, we define $r=\tilde{r}^{\frac{n_s}{n_m}}$, then $\hat{D}(\|x-z\|)=\tilde{r}$ is equivalent to $\|x-z\|=r$.  Hence
\begin{equation}\label{eq:density}
\rho( \tilde{r}; \mathcal{M},f, \hat{D})d \tilde{r} = p_{x,z\sim f(\mathcal{M})}(\hat{D}=\tilde{r})d\tilde{r} = p_{x,z\sim f(\mathcal{M})}(\|x-z\|=r) dr
\end{equation}
By Assumption 2, this last term is
\[
p_{x,y\sim f(\mathcal{M})}(\|x-z\|=r)= \rho( r; \mathcal{M},f, \|\cdot\|) = cn_m r^{n_m-1}
\]
Inserting this into \eqref{eq:density} we obtain 
\[
\rho( \tilde{r}; \mathcal{M},f, \hat{D})=  p_{x,y\sim f(\mathcal{M})}(\|x-z\|=r)\cdot \frac{dr}{d\tilde{r}} = cn_m r^{n_m-1}\cdot \frac{dr}{d\tilde{r}} = cn_m r^{n_m-1}\cdot \frac{n_s}{n_m} \tilde{r}^{\frac{n_s}{n_m}-1} = cn_s \tilde{r}^{n_s-1}
\]
Hence we proved \eqref{eq:toshow}.
\end{proof}

\bibliographystyle{siamplain}
\bibliography{references}

\end{document}